# Offensive Language Detection in Under-resourced Algerian Dialectal Arabic Language

Oussama Boucherit [1] and Kheireddine Abainia[2]

PIMIS Laboratory, Departement of Electronics & Telecommunications,
Université 8 Mai 1945, Guelma, 24000, Algeria
[1]`boucherit.oussama.24@gmail.com`
[2]`k.abainia@univ-guelma.dz`

**Abstract.** This paper addresses the problem of detecting the offensive and abusive content in Facebook comments, where we focus on the Algerian dialectal Arabic which is one of under-resourced languages. The latter has a variety of dialects mixed with different languages (i.e. Berber, French and English). In addition, we deal with texts written in both Arabic and Roman scripts (i.e. Arabizi). Due to the scarcity of works on the same language, we have built a new corpus regrouping more than 8.7k texts manually annotated as normal, abusive and offensive. We have conducted a series of experiments using the state-of-the-art classifiers of text categorisation, namely: BiLSTM, CNN, FastText, SVM and NB. The results showed acceptable performances, but the problem requires further investigation on linguistic features to increase the identification accuracy.

## 1 Introduction

Due to the rapid development of social media, online communication becomes more easier. However, this rises some concerns such as the use of offensive language that may harm people mentally and physically. Offensive language can take various forms including hate speech, bullying, disrespect, abuse and violence. This behavior may lead to depression which has a negative impact on people's health and relationships, and it may lead to suicide as well. To overcome this issue, several researchers worked on the automatic detection of such kind of content in a myriad of languages of the world.

Serval Arabic studies have been carried out to detect offensive language, but in social media people often use dialectal Arabic which differs from official Arabic. There is a scarcity of research works undergone on the Algerian Arabic, because the latter is known with a complicated linguistic structure and contains several terms borrowed from different languages such as French, Italian, Spanish, and Turkish. All these issues among others make automatic processing of such kind of texts more difficult.

In this paper, we investigate automatic offensive language detection in under-resourced Algerian dialectal Arabic, where we propose a new corpus (i.e. DziriOFN) for this task due to the lack of works on the same dialectal Arabic. The created corpus has been annotated by five native speakers at the sentence level, wherein the texts are labelled as offensive, abusive and normal. In addition, we have evaluated the state-of-the-art tools of text categorization such as SVM, NB, CNN, BiLSTM and FastText. The experimental results showed that the conventional tools produce acceptable results, but they require further investigation to enhance the accuracy.

## 2   Related work

In this section, we highlight some research works carried out on the offensive language detection, where we state related works in Latin and Arabic languages.

For Latin languages, a statistical classifier based on sentiment analysis has been proposed by Gitari, et al. (2015), wherein the authors proposed a model to detect the subjectivity. The proposed model did not only detect if the sentence is subjective, but it rather can identify and score the polarity of sentiment expressions. Another statistical classifier has been proposed, for which the authors collected over 1M Tweets to detect hate speech in Italian language (Santucci et al., 2018).

A machine learning approach based on feature selection of meta-data has been proposed to deal with automatic offensive language detection in Twitter (De Souza and Da Costa-Abreu, 2020). In particular, the authors used SVM and NB classifier, where the latter outperformed the SVM (i.e. 92% of accuracy in contrast to 90%). Another classifier combining SVM and MLP (Multilayer perceptron) has been proposed in (Nayel and Shashirekha, 2019), where the authors used Stochastic Gradient Descent (SGD) for feature selection. They experimented with three Indo-European Languages (i.e. English, German and Hindi), and the results showed that the proposed framework was suitable with the Hindi language rather than English and German. Greek offensive language identification has been proposed, wherein the authors created a new Twitter corpus (OGTD) containing 4.7K texts manually annotated (Pitenis et al., 2020). The authors tested different ML and DL approaches such as SVM, SGD, NB and LSTM. The experimental results on OGTD showed that LSTM (with Attention) outperformed the conventional ML approaches (i.e. 0.89 of F1-score).

A hate speech detection approach established a lexical baseline for discriminating between hate speech and profanity on a standard dataset (Zampieri et al., 2019). In (Badjatiya et al., 2017) a Twitter corpus of 16K texts for hate speech identification was manually annotated, and for which various DL approaches have been evaluated. Another research using DL approaches has been proposed for hate speech detection in Indonesian Language (Sutejo and Lestari, 2018). The authors evaluated different feature models, where textual features produced promising results (87.98% of F1-score).

Some studies have been carried out in Arabic language. For instance, an approach for detecting cyberbullying in Arabic texts has been proposed by Haidar et al. (2017), where the approach focused on preventing cyberbullying attacks. In particular, it uses NLP techniques to identify and process Arabic words, and ML classifiers to detect the bullying content.

A dataset for Arabic hate speech detection with 9.3k annotated tweets was proposed by Raghad and Al-Khalifa (Raghad and Al-Khalifa, 2020). The authors experimented with several DL and ML models to detect hate speech in Arabic tweets. The results showed that CNN-GRU produced the best performances (0.79 of F1-score). A multitask approach for Arabic offensive language and hate-speech detection using DL, transfer learning and multitask learning was proposed in (Abu-Farha and Magdy, 2020). Otiefy et al. (2020) experimented the offensive language identification on multiple Twitter datasets, where the authors evaluated several ML and DL models. They used a combination of characters and words n-gram in a linear SVM model, which produced the best performance among the other baseline models (Otiefy et al., 2020).

A multitask approach to detect Arabic offensive language has been proposed by Djandji et al. (2020), where the proposal could be trained with a small training set. The proposed approach ranked the second among others in both tasks of the shared task (i.e 90% and 82.28% of F1-score). Solving the problem of out-of-vocabulary in Arabic language has been proposed for detecting the offensive language, where the authors presented a model with character-level embeddings (Alharbi and Lee, 2020).

An approach for abusive language detection on Arabic social media (dialectal Arabic) has been proposed, where two datasets were introduced (Mubarak et al., 2017). The first contains 1,100 manually labeled dialectal tweets, and the second contains 32K comments that the moderators of popular Arabic newswires deemed inappropriate. The authors have proposed a statistical approach based on a list of obscene words, and the produced results were around 60% of F1-score. Hate speech and abusive detection approach in Levantine dialect has been proposed by Mulki et al. (2019), where the authors experimented with both binary classification and multi-class classification employing SVM and NB classifiers.

Different ML approaches and an ensemble classifier have been used to deal with offensive language identification in dialectal Arabic (Husain, 2020). The conducted study showed an interesting impact of the preprocessing on such task, as well as good performances of the ensemble classifier in contrast to standard ML algorithms. Another work focused on Tunisian hate speech and abusive speech has been proposed, in order to create a benchmarked dataset (6k of tweets) of online Tunisian toxic contents (Haddad et al., 2020). The authors evaluated two ML approaches (i.e. NB and SVM), where the NB classifier outperformed the SVM (92.9% of accuracy).

## 3   Corpus

To the best of our knowledge, there is only one similar corpus proposed for offensive language (i.e. hate speech) on Algerian dialectal Arabic (Guellil et al., 2021). This corpus was proposed for hate speech detection against women, and crawled from Youtube social media. Unfortunately, the corpus regroups only 3.8k texts labelled as "*not hateful*" and "*hateful*", where the latter has 792 compiled texts. We think that this is not enough to train machine learning classifiers to correctly recognize hateful texts, because we cannot cover different writing possibilites (i.e. dialectal texts).

We have created a new[1] (i.e. DziriOFN) for the same task, but it is for the offense detection in general and not addressed to specific target (e.g. women). Our corpus was crawled from Facebook social media, while the latter is considered as the first communication media used by the Algerian community.

### 3.1 Data collection

Firstly, we have selected a set of public pages and groups related to sports and politics. Among the posts, we have selected the ones addressing harmful and provoking subjects that involve more interactions. More specifically, the most of the subjects contain conflicts and controversies in news, religion, ethnecity and football. The Algerian community is conservative and some provoking subjects may involve offensiveness while expressing opinions.

**Table 1.** Details of the annotation rounds

|             | 1st round | 2nd round | 3rd round |
|-------------|-----------|-----------|-----------|
| Annotator #1 | 6,000    | 0         | 4,258     |
| Annotator #2 | 4,258    | 0         | 6,000     |
| Annotator #3 | 0        | 3,000     | 0         |
| Annotator #4 | 0        | 3,000     | 0         |
| Annotator #5 | 0        | 4,258     | 0         |

Because of Facebook politics, public users are restricted to gather public or private data from this platform. In this regard, three Javascript scripts[2] were created to automate the data collection instead of doing the task manually. The scripts unhide all the comments of a given post, unhide the second part of long comments and retrieve all the comments with their informations (i.e. user names, profile links, comment texts, number of reactions and number of replies).

Empty comments or comments with only images and emojies were ignored. In overall, we have crawled 10,258 comments written in Arabic script, Roman script (i.e. Arabizi) or both.

### 3.2 Data annotation

To annotate our corpus, five Algerian native speakers were involved in this task using an in-house crowdsourcing platform. They were instructed to attribute one label to each text among three labels, i.e. offensive, abusive or normal, as well as one language label among three labels, i.e. Arabic (MSA), dialect or mixed between Arabic and dialect. The text is ignored if it is not offensive and completly written in another language (French, English or Berber), or it is unclear (not understood) or difficult to label.

We have defined offensive texts as texts that contain hateness, agressiveness, bullying, harassment, violence or offend a target (someone, a group of people or an entity in general). On the other hand, we have defined abusive texts as texts containing swear words or sexual/adult content.

---

[1] https://github.com/xprogramer/DziriOFN
[2] https://github.com/xprogramer/fb-cmt-crawl

**Table 2.** Corpus description

|  | Number of texts |
|---|---|
| Offensive texts | 3,227 |
| Abusive texts | 1,334 |
| Normal texts | 4,188 |
| Ignored texts | 1,509 |
| *Total* | *10,258* |

The annotation was performed in three rounds. In the first one, annotator #1 annotated the first 6,000 texts and annotator #2 annotated the remaining 4,258 texts. In the second round, annotator #3 annotated the first 3,000 texts, annotator #4 annotated the following 3,000 texts and annotator #5 annotated the remaining 4,258 texts. Finally, in the third round, if there is a conflict between the two first rounds, annotator #2 annotated the first 6,000 texts (only conflicted labels) and annotator #1 annotated the remaining 4,258 texts (only conflicted labels).

In overall, from the collected data, 1,509 texts were ignored (for various reasons), 3,227 texts were labelled as offensive, 1,334 texts were labelled as abusive and 4,188 texts were labelled as normal (Table 2).

**Table 3.** Inter-annotator agreement between different annotators

| Category | $1^{st}$ vs $2^{nd}$ | $1^{st}$ vs $3^{rd}$ | $2^{nd}$ vs $3^{rd}$ |
|---|---|---|---|
| Normal | 3,449 | 3,603 | 4,008 |
| Abusive | 1,108 | 85 | 141 |
| Offensive | 2,341 | 583 | 305 |

**Table 4.** Labels overlaping between final labels and annotator labels

| Category | $1^{st}$ annotator | $2^{nd}$ annotator | $3^{rd}$ annotator |
|---|---|---|---|
| Normal | 3,635 | 4,021 | 4,175 |
| Abusive | 1,199 | 1,249 | 226 |
| Offensive | 2,944 | 2,645 | 887 |

Table 3 summarizes the inter-annotator agreements between different annotators, where the third annotator refers to the ensemble of annotators (i.e. 4th, 5th and 6th). It is noticed that normal texts present a high level of agreement between different annotators, because it is more easier to differentiate such kind of texts than the other categories. In addition, the first and the second annotator agreed in the most of the abusive and offensive texts (around 83% and 72.5%, respectively).

By comparing Table 3 and Table 4, the third annotator correctly labelled more normal texts in contrast to the two others. Conversely, the third annotator (i.e. 4th, 5th and 6th) did not correctly label the offensive and abusive texts (around 27.5% and 16.9%), while annotator #1 correctly labelled the most of abusive texts (around 91.2%) and annotator #2 correctly labelled the most of offensive texts (around 93.6%).

Thus, annotator #3 highly contributed in the annotation of normal texts, while annotator #1 and #2 contributed in the annotation of abusive and offensive texts (and normal texts as well). It is obvious that the data annotation depends on the gender, age and culture of the annotators, and maybe the set of the third annotator are familiar with offensive texts and considered them as normal texts. It is also noticed that some abusive texts were labelled as offensive by the set of the third annotator, and this is due to the lack of training (not well trained) or they considered them as as offensive regarding their beliefs.

## 4    Experimental results

We have evaluated some ML and DL classifiers on our corpus. For machine learning, we have used SVM, MultinomialNB and GaussianNB with default settings of scikit-learn toolkit. We have applied standard preprocessing steps (i.e. removing punctuation marks), and applied the tf-idf technique to weight the word frequencies.

For deep learning classifiers, we have used 512 filters with [3 4 5] as filter size and 0.5 of a dropout rate in CNN. Similarly, we have built the BiLSTM model with one hidden layer and 0.5 of dropout. However, we have used the default settings of FastText (Bojanowski et al., 2017).

In overall, we have conducted two series of experiments, i.e. binary classification and multi-class classification. In the first one, we merged offensive and abusive comments together, while in the second experiment we considered each category independently (i.e. three classes). The corpus was split into training set and test set (90% and 10%, respectively).

**Table 5.** Test results of ML models trained with two and three labels.

|  | **SVM** | **Multinomial NB** | **Gaussian NB** |
|---|---|---|---|
| **Two labels** | 0.744 | 0.752 | 0.710 |
| **Three labels** | 0.669 | 0.662 | 0.518 |

The results of Table 5 summarize the accuracies produced by SVM, MultinomialNB and GaussianNB trained with two classes and three classes. Both SVM and Multinomial NB reported high accuracies in both experiments, while Gaussian NB was the worst and considerably decreased in three-label classification.

From Table 6, deep learning classifiers produce low performances compared to machine learning classifiers. However, FastText highly outperformed CNN and BiLSTM, but produced slightly reduced accuracy compared with Multinomial NB and SVM. Moreover, deep learning classifiers in binary classification perform better than the multi-label classification (three classes).

It is obvious that as more the number of classes decreases as more the identification accuracy increases, and vice-versa. The reason of low performances produced by DL classifiers is that such classifiers require a huge training set, as well as they cannot well extract the features in the lack of standard orthography (different writing possibilites). Indeed, as we experiment with words uni-gram, we cannot cover all writing variants of the words, but maybe with character n-grams the accuracy may increase, because wathever the words writing is chaning they keep some common characters (generally vowels change). In addition, sometimes it is difficult to spot abusive and offensive texts even by the humans in the lack of strongly offensive words, and that is why automatic algorithms sometimes cannot differentiate between normal and offensive texts.

**Table 6.** Test results of DL models trained with two and three labels.

| Category | CNN | BiLSTM | FastText |
|---|---|---|---|
| **Two labels** | 0.523 | 0.520 | 0.716 |
| **Three labels** | 0.347 | 0.400 | 0.648 |

Finally, SVM and Multinomial NB hold the best performances among the others, because the words have been weighted with td-idf that gives low frequencies to stop words and common words used in any text category, while unique words (or terms) have high frequencies.

## 5   Conclusion

In this work, we have addressed the problem of offensive language identification in under-resourced Algerian dialectal Arabic, for wich we have created a new corpus for the addressed problem because of the scarcity of works carried out on the same language. The corpus was crawled from Facebook social media (common used social media in Algeria), where 10,258 comments have been gathered from public pages and groups related to sensitive topics. Five annotators were involved in the annotation task by following a general guideline, and the texts were labeled into one of the three categories, i.e. offensive, abusive and normal. We have evaluated the state-of-the-art tools of text categorization such as SVM, Miltinomial NB, Gaussian NB, CNN, BiLSTM and FastText, where we have carried out two sets of experiments, i.e. binary classification and multi-label classification. In the first one, we have merged offensive and abusive texts in the same category (offensive), while in the second experiment we have kept the categories independently.

The experimental results showed that SVM and Multinomial NB classifiers outperformed all the other classifiers in both experiments (binary and multi-label classification). The results were acceptable, but the algorithms require further investigation to improve the accuracy, because word uni-grams cannot cover all different writing possibilites.

As a future work, we plan to build a larger corpus while enhancing and adding new rules in the preprocessing. Moreover, we plan to propose a new algorithm based on rules to detect offensive and abusive language effectively.